\documentclass[runningheads]{llncs}
\pdfoutput=1
\usepackage{eccv}

\usepackage{eccvabbrv}

\usepackage{graphicx}
\usepackage{booktabs}

\usepackage[accsupp]{axessibility}  % Improves PDF readability for those with disabilities.
\usepackage{placeins}
\usepackage{afterpage}

\usepackage{hyperref}

\usepackage{orcidlink}
\usepackage[bb=boondox]{mathalfa}
\usepackage{amsmath}
\usepackage{multirow}
\usepackage{algpseudocode}
\usepackage{algorithm}
\usepackage{tikz}
\usetikzlibrary{shapes, arrows}

\DeclareMathOperator{\diag}{diag}

\begin{document}

\title{SINDER: Repairing the Singular Defects of DINOv2}

\titlerunning{SINDER}

\author{Haoqi Wang\inst{1}\orcidlink{0000-0002-5760-4097} \and
  Tong Zhang\inst{1}\orcidlink{0000-0001-5818-4285} \and
  Mathieu Salzmann\inst{1,2}\orcidlink{0000-0002-8347-8637}}

\authorrunning{H.~Wang et al.}

\institute{School of Computer and Communication Sciences, EPFL, Switzerland
   \and
   Swiss Data Science Center, Switzerland \\
  \email{\{haoqi.wang,\;tong.zhang,\;mathieu.salzmann\}@epfl.ch}
}

\maketitle

\begin{abstract}
  Vision Transformer models trained on large-scale datasets, although effective, often exhibit artifacts in the patch token they extract.
While such defects can be alleviated by re-training the entire model with additional classification tokens, the underlying reasons for the presence of these tokens remain unclear.
In this paper, we conduct a thorough investigation of this phenomenon, combining theoretical analysis with empirical observations.
Our findings reveal that these artifacts originate from the pre-trained network itself, specifically stemming from the leading left singular vector of the network's weights.
Furthermore, to mitigate these defects, we propose a novel fine-tuning smooth regularization that rectifies structural deficiencies using only a small dataset, thereby avoiding the need for complete re-training.
We validate our method on various downstream tasks, including unsupervised segmentation, classification, supervised segmentation, and depth estimation, demonstrating its effectiveness in improving model performance.
Codes and checkpoints are available at \url{https://github.com/haoqiwang/sinder}.

\end{abstract}

\section{Introduction}\label{sec:intro}

\begin{figure}[tb]
  \centering
  \includegraphics[height=4.8cm]{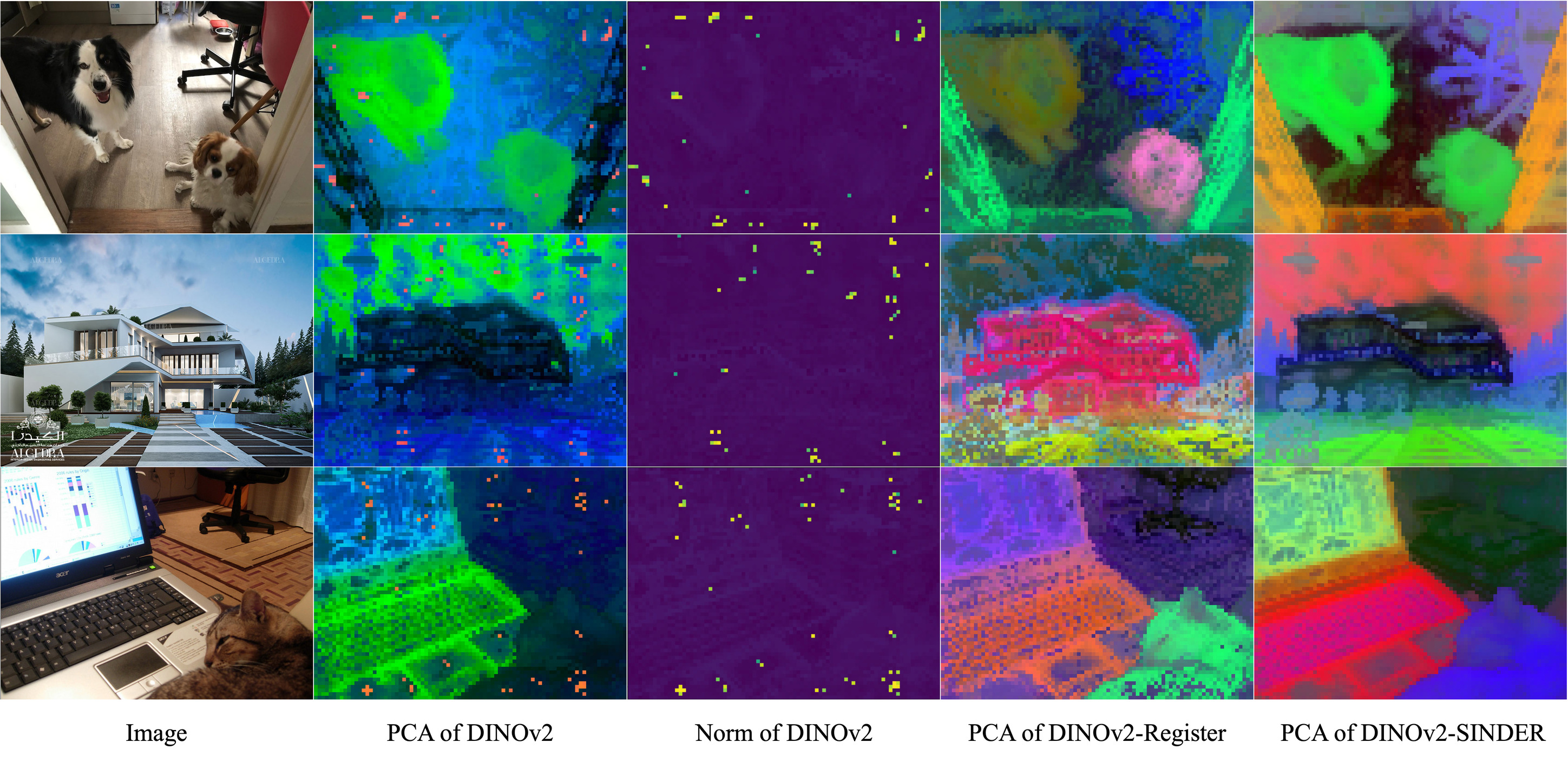}
  \caption{Visualization of singular defects in the feature map of the last layer of DINOv2.
    The images are resized to have height \(896\) when input into the networks.
    The color of the PCA visualization comes from the three principal components of the patch tokens.
  }\label{fig:high_norm}
\end{figure}

Self-supervised learning (SSL) has emerged as a highly effective method for network pre-training~\cite{oquab2023dinov2}, producing features beneficial across a wide spectrum of downstream tasks.
SSL significantly accelerates large-scale training for vision models, exemplified by recent advancements such as DINOv2~\cite{oquab2023dinov2}.
While SSL models excel in image classification tasks, their use for comprehensive image understanding, e.g., segmentation, is significantly challenged by the presence of defective patch tokens, as depicted in Figure~\ref{fig:high_norm}.
These anomalies materialize as high-norm tokens in the feature maps of vision transformers. Despite research efforts to understand this phenomenon~\cite{darcet2023vitneedreg}, explanations remain in their infancy.
Current observations indicate that these flawed patches offer minimal local information, suggesting a tendency for large and deep vision transformers to recycle redundant patch tokens to store more useful information. To this day, the only approach to addressing this issue~\cite{darcet2023vitneedreg} requires re-training the network from scratch with additional register tokens on vast amounts of data, which is typically prohibitive and offers limited explanations of the underlying phenomenon.

In this paper, we aim to bridge the understanding gap of these defects by providing mathematical explanations.
In contrast to previous work attributing the defects to the image classification token, we discover that these defects inherently exist and share high similarity across the entire dataset.
To further explore and develop theoretical foundations, we linearize the weights of each network block. Our analysis reveals a strong correlation between the defects in each layer and the corresponding leading left singular vector of the linearized operations. We thus term this phenomenon \emph{singular defects}.
Importantly, our analysis evidences that such singular defects depend solely on the pre-trained network weights, and not on the inputs.

To mitigate these singular defects, we propose a method based on fine-tuning the singular values of linear layers in the network.
Specifically, we impose a smoothness regularization on the detected defective tokens to rectify them and restrict the number of learnable parameters to as few as possible to retain the original feature quality.
Our approach can fine-tune pre-trained large-scale models using only a small dataset without the need for labeled data.
Our experimental results demonstrate that our method effectively rectifies these defects and enhances performance in semantic segmentation tasks, particularly in the unsupervised setting, while maintaining performance in classification tasks. Compared to the existing solution of~\cite{darcet2023vitneedreg}, which requires retraining networks on private LVD-142M data~\cite{oquab2023dinov2}, our method offers advantages in terms of reduced carbon emissions, memory footprint, and time consumption.
Considering the limited availability of proprietary large-scale datasets, our approach offers a viable and economical-friendly solution for deploying large-scale models.

In a nutshell, our contributions can be summarized as follows:
\begin{itemize}
  \item We unveil the correlation between the leading left singular vector and the direction of defects, enabling us to theoretically predict the direction of defects for each layer.
  \item  We introduce a data-efficient fine-tuning technique to address the singular defects of DINOv2 without necessitating access to large-scale datasets.
  \item  We conduct extensive experiments to study the properties of the defects and show that our proposed solution not only retains the feature quality for downstream classification tasks but also improves the pixel-level prediction tasks such as unsupervised segmentation.
\end{itemize}

\section{Motivation}
A prominent characteristic of the defective patch tokens in the last layer is their high norm, a phenomenon that was also observed in previous work~\cite{darcet2023vitneedreg}.
We illustrate the norms of the patch tokens of several images in Figure~\ref{fig:high_norm}.
The norm of the defective tokens is much larger than that of normal tokens.
For example, on 500 randomly selected natural images from the validation set of ImageNet-1K, their average norms are 434.0 vs. 57.6, respectively.

The second characteristic of defective patch tokens is that their directions are largely image-independent.
Firstly, the feature directions of the defective patch tokens within each image are almost the same.
This can be intuitively seen from Figure~\ref{fig:high_norm}, whose second column depicts a PCA representation of the tokens.
High-norm tokens in the same image have the same color, indicating that their directions are close to each other.
To quantify this, we compute the average pair-wise angles between the defects within each of the 500 images.
Their mean is 3.1 degrees.
By contrast, the statistics of the average pair-wise angles between all patch tokens within each image is 72.8 degrees.

Secondly, the defective patch tokens are almost the same across different images.
To see this, we calculate the average defective tokens for each image and then compute the average pair-wise angles between them.
The mean is 5.5 degrees.
This confirms that the defect direction in the last layer is in essence input image agnostic.
This observation differs from that in~\cite{darcet2023vitneedreg}, where high-norm tokens were claimed to contain image-wise global information.
Based on our statistics, these high-norm tokens primarily contain no input information, regardless of local or global.
It thus seems natural to ask, \emph{can the defect directions be directly inferred from the pre-trained weights without knowing the input image?}

These two observations of defects remind us of the power method~\cite{wiki:Power_iteration} in linear algebra, where a vector is recursively multiplied by a square matrix;
the vector converges to the leading eigenvector, regardless of its initial direction, and its norm explodes if the largest eigenvalue is larger than 1.
This motivates us to approach the problem of defects from the perspective of singular value decomposition, which will be the central topic of Section~\ref{sec:method}.

As a sanity check, we clamp the singular values of the weights of all the linear layers in DINOv2 to a smaller value and found that the high-norm defects are reduced (visualizations can be found in the Appendix).
Encouraged by this, in Section~\ref{sec:repair}, we design a regularization strategy to limit the magnitude of the singular values and thus the defective patch tokens.

\section{Singular Defect Direction}\label{sec:method}

Now we delve into the origin of singular defects, focusing on the DINOv2 giant model.
The DINOv2 giant is a Vision Transformer (ViT) model comprising 40 transformer layers, each containing an Attention Block and an MLP Block.
These blocks act as residuals, with an identity path connecting their input and output.
Our objective is to analyze the influence of each block on the defective tokens and to predict defective token directions theoretically solely from the pre-trained weights of the network, in an input-agnostic manner.

To evaluate the quality of our theoretical predictions, we manually extract the defect directions of 500 images from the ImageNet validation set.
For each layer, we compute the average defect direction across these images, termed the \emph{empirical defect direction}.
Since evident defects on feature maps only manifest after the 15th layer, we focus solely on gathering defect directions from the 15th layer onwards.
A good theoretical estimation of the defect direction is expected to closely align with this empirical defect direction.

It's worth noting that to facilitate tractability in our analysis, we consider the simplified scenario where there is \emph{only one input token}.
Under this assumption, the transformer layer can be approximated by linear transformations, as we will demonstrate below, rendering theoretical analysis feasible.
Although the analysis is conducted in a simplified setting, we confirmed that the theoretical predictions of the singular defects are accurate.

\tikzstyle{terminator} = [rectangle, draw, text centered, rounded corners, minimum height=2em]
\tikzstyle{starter} = [text centered, rounded corners, minimum height=2em]
\tikzstyle{connector} = [draw, -latex']
\subsection{Linear Approximation of an Attention Block}
For an input token \(x \in \mathbb{R}^D\), the computation of the Attention Block is \\
{\small \noindent
\begin{tikzpicture}
  \node [starter,] at (0,0.2) () {};
  \node [starter,] at (0,-0.1) () {};
  \node [starter, fill=red!20] at (0,0) (x) {$~x~$};
  \node [terminator, fill=blue!20] at (1.8,0) (layernorm) {\texttt{layer norm}};
  \node [terminator, fill=blue!20] at (5.25,0) (attention) {\texttt{multi-head attention}};
  \node [terminator, fill=blue!20] at (8.85,0) (layerscale) {\texttt{layer scale}};
  \node [starter, fill=green!20] at (11.15,0) (output) {output.};
  \path [connector] (x) -- (layernorm);
  \path [connector] (layernorm) -- (attention);
  \path [connector] (attention) -- (layerscale);
  \path [connector] (layerscale) -- (output);
\end{tikzpicture}} \\
Under the single-token assumption, we will show that we can approximate these operations as linear transformations.
Let us first study the \texttt{layer norm}. This operation is non-linear because of the division by the standard deviation.
However, if we ignore this rescaling, the rest is linear and can be written as \(A_1(A_0x)+b_1\),
where \(A_0=I-\frac{1}{D}\mathbb{1}_{D\times D}\) is the centering, \(A_1=\diag(w)\) is the diagonal matrix of scaling parameters, and \(b_1\) contains the bias parameters.

For the \texttt{multi-head attention}, as we analyze for a single token case, the softmax over a singleton is a constant 1. Hence, we only need to consider the \texttt{value} parameters.
Let the weights concatenated over all \texttt{value} heads be \(A_2\in \mathbb{R}^{D\times D}\), the concatenated biases from all heads be \(b_2\in \mathbb{R}^D\), and the weights and biases of the output projection be \(A_3\) and \(b_3\), respectively, then the multi-head attention can be written as \(A_3(A_2x+b_2)+b_3\).

Finally, we rewrite the \texttt{layer scale}  as \(A_4x\), where \(A_4=\diag(w)\) is the diagonal matrix of scaling parameters.

Combining the above operations, the Attention Block can be approximated as a series of linear transformations,
\begin{equation}
  \mathrm{Attention}(x) \approx A_4(A_3(A_2(A_1(A_0x)+b_1)+b_2)+b_3):=Ax+b.
\end{equation}

\begin{figure}[t]
  \centering
  \begin{subfigure}{0.49\linewidth}
    \includegraphics[width=0.99\linewidth]{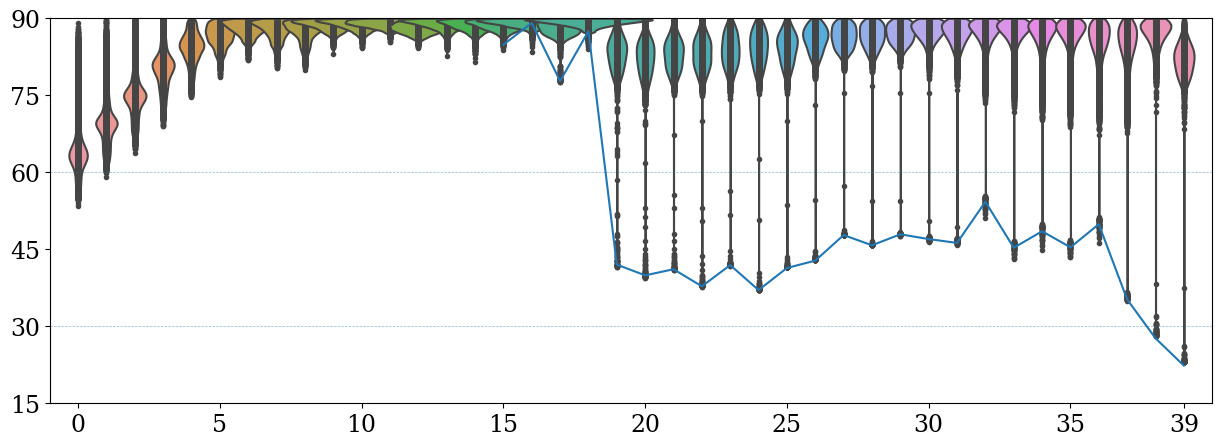}
    \caption{Attention Block with Identity \(I+A\)}\label{fig:theoretical-a}
  \end{subfigure}
  \begin{subfigure}{0.49\linewidth}
    \includegraphics[width=0.99\linewidth]{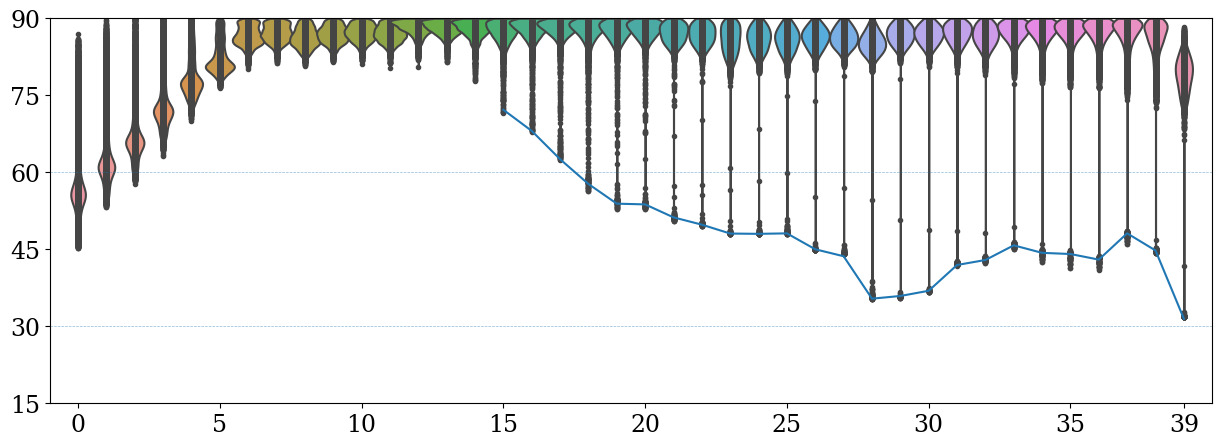}
    \caption{MLP Block with Identity \(I+C\)}\label{fig:theoretical-b}
  \end{subfigure}
  \begin{subfigure}{0.49\linewidth}
    \includegraphics[width=0.99\linewidth]{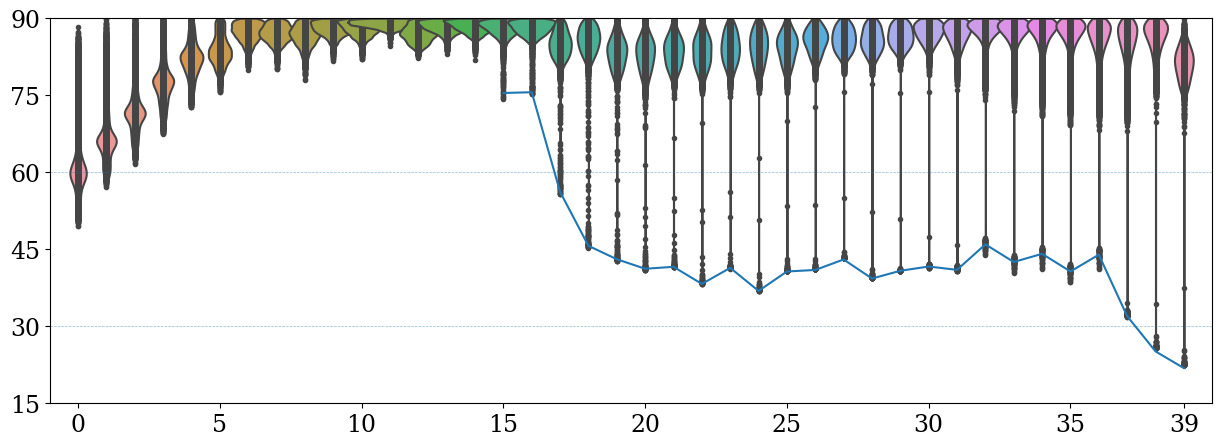}
    \caption{Layer: Attention and MLP Block \(E\)}\label{fig:theoretical-c}
  \end{subfigure}
  \begin{subfigure}{0.49\linewidth}
    \includegraphics[width=0.99\linewidth]{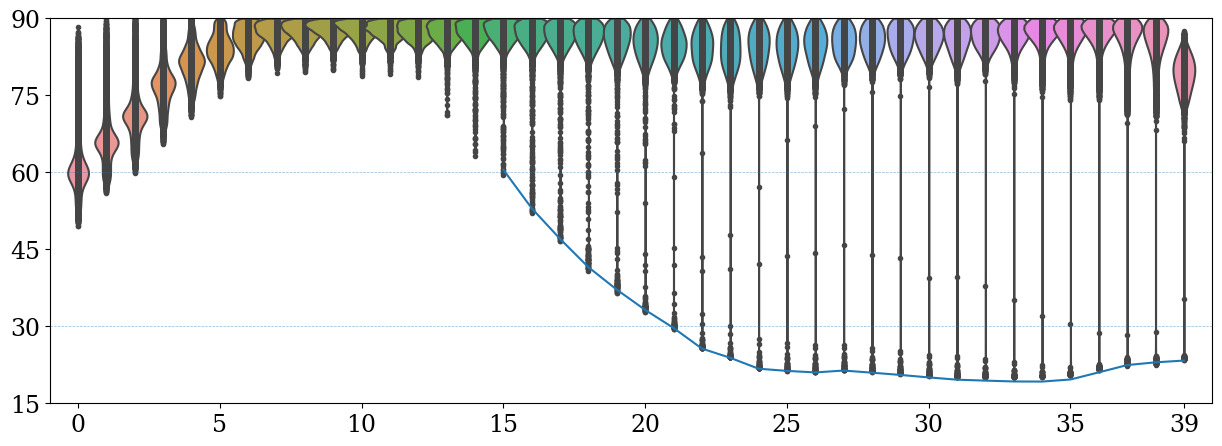}
    \caption{Composed multiple layers \(G\)}\label{fig:theoretical-d}
  \end{subfigure}
  \caption{{\bf Angle between theoretical and empirical defect directions.} Blue lines are the angle between the empirical defect direction and the leading left singular vector of \(I+A\), \(I+C\), \(E\), \(G\), respectively.
    Angles between the leading left singular vectors and all the patch tokens in each layer are shown as violin plots.
    The \(x\)-axis is the layer index, and the \(y\)-axis is the acute angle in degrees.
    The villa image in Figure~\ref{fig:high_norm} is used.}\label{fig:theoretical}
\end{figure}

Drawing inspiration from the power method, we relate the empirical defect directions with the leading left singular vector corresponding to the largest singular value of \(I+A\).
In Figure~\ref{fig:theoretical-a}, we depict the angles between the leading left singular vector and the empirical defective tokens for each layer as a blue line.
It is evident that after layer 19, these angles converge, with values consistently below 40 degrees.
Additionally, we present the angles between the leading left singular vector and all patch tokens of an image as violin plots.
The defective tokens are identifiable as isolated points within the violin plots (more visualizations in the Appendix).
These findings suggest that the angles between the leading left singular vector and the patch tokens serve as a reliable metric for detecting defective tokens.

\afterpage{\FloatBarrier}

\subsection{Linear Approximation of an MLP Block}
The computation graph for an MLP Block is\\
{\small \noindent
\begin{tikzpicture}
  \node [starter,] at (0,0.2) () {};
  \node [starter,] at (0,-0.1) () {};
  \node [starter, fill=red!20] at (2,0) (x) {$~x~$};
  \node [terminator, fill=blue!20] at (3.8,0) (layernorm) {\texttt{layer norm}};
  \node [terminator, fill=blue!20] at (5.7,0) (mlp) {\texttt{mlp}};
  \node [terminator, fill=blue!20] at (7.7,0) (layerscale) {\texttt{layer scale}};
  \node [starter, fill=green!20] at (10,0) (output) {output,};
  \path [connector] (x) -- (layernorm);
  \path [connector] (layernorm) -- (mlp);
  \path [connector] (mlp) -- (layerscale);
  \path [connector] (layerscale) -- (output);
\end{tikzpicture}} \\
where \texttt{layer norm} and \texttt{layer scale} can be processed in the same manner as in the Attention Block.
However, the \texttt{mlp} layer is non-linear, expressed as \(C_3(\mathrm{silu}(W_1x+h_1)\odot (W_2x+h_2))+d_3\), where \(\odot\) is the element-wise product, \(C_3\), \(W_1\), \(W_2\) are weights, and \(h_1\), \(h_2\), \(d_3\) are the biases.
We approximate the \texttt{mlp} using least squares.
Specifically, we sample 100,000 random inputs \(X\in \mathbb{R}^{D\times 100,000}\) and compute their outputs \(Y=\mathrm{silu}(W_1X+h_1)\odot (W_2X+h_2)\in \mathbb{R}^{M\times 100,000}\), where \(M\) is the output dimension of \(W_1\).
We solve the least-square problem \(C_2X=Y\), and obtain the linear approximation with matrix \(C_2\in \mathbb{R}^{M\times D} \).

Ultimately, we can approximate the MLP Block as
\begin{equation}
  \mathrm{MLP}(x) \approx C_4(C_3(C_2(C_1(C_0x)+d_1))+d_3):=Cx+d,
\end{equation}
where \(C_0=I-\frac{1}{D}\mathbb{1}_{D\times D}\) is the centering of the \texttt{layer norm}, \(C_1\) is the diagonal matrix of the  \texttt{layer norm} scaling weights, \(d_1\) is the bias of the \texttt{layer norm}, and \(C_4\) is the diagonal matrix of the \texttt{layer scale} scaling weights.
The angles between the leading left singular vector of \(I+C\) and the empirical defect directions are shown in Figure~\ref{fig:theoretical-b}.

\subsection{Combining Attention and MLP Blocks}
Based on our previous approximations, we can combine the linearized Attention Block and MLP Block as follows,
\begin{equation}
  \mathrm{Layer}(x) \approx x+Ax+b+C(x+Ax+b)+d := Ex + f,
\end{equation}
where the identity path is incorporated.
We plot the effect of the leading left singular vector of \(E\) in Figure~\ref{fig:theoretical-c}, which resembles both Figure~\ref{fig:theoretical-a} and Figure~\ref{fig:theoretical-b}.

\subsection{Predicting Defective Token Direction for Each Layer}\label{sec:theoretical}
We can further improve the prediction of defect direction by composing the linear approximations from layer \(0\) to layer \(i\), where the matrix multiplied with \(x\) is
\begin{equation}
  G_i := E_iE_{i-1}\cdots E_0.
\end{equation}
The result of the leading left singular vector of \(G_i\) is shown in Figure~\ref{fig:theoretical-d}.
We find that after layer 20, the leading left singular vectors are very close to the empirical defect directions, and from layers 15 to 19, the result is also better than previous attempts.
Thus, we define the leading left singular vector of \(G_i\) as the theoretical \emph{singular defect direction}\footnote{We do not differentiate between a singular defect direction and its negative direction.} for layer \(i\).
Figure~\ref{fig:theoretical-d} demonstrates that we can accurately predict the empirical defect direction by the singular defect direction.
So we had referred to this type of defective tokens as \emph{singular defects}.

Note that the definition of singular defect direction originates solely from the pre-trained network weights; it does not depend on the input image in inference.

\section{Repairing Singular Defects}\label{sec:repair}
Having identified the connection between the singular defect direction and the empirical defect direction, we next aim to repair the singular defects of the network with minimum modifications to the network parameters.
A key requirement of such repairs is that they should maintain the feature quality of the original network.
Without the defective tokens, we expect a spatially smooth and coherent feature map, thus leading to stronger performance for dense prediction downstream tasks.
We identify two key aspects that contribute to this goal.
First, imposing smooth regularization suffices to ensure the resulting network produces spatially smooth and coherent feature maps.
Second, to maintain the feature quality, the algorithm should modify as few parameters as possible, refraining from overtraining the network.

Based on these observations, we design an algorithm called \emph{\textbf{Sin}gular \textbf{De}fect \textbf{R}epairing} (SINDER, Algorithm~\ref{alg:algorithm}).
In essence, SINDER aims to repair the first defective layer encountered in the forward pass using a smooth regularization, by modifying a few parameters.
We describe the corresponding loss in Section~\ref{sec:loss} and discuss the importance of limiting the number of learnable parameters in Section~\ref{sec:limit}.

\renewcommand{\algorithmicrequire}{\textbf{Input:}}
\begin{algorithm} \caption{Singular Defect Repairing (SINDER)}\label{alg:algorithm}
    \begin{algorithmic}[1]
        \Require{}A pre-trained network, a finetune dataset, termination threshold \(\rho=25\%\), \(M=500\), skip threshold \(\sigma=3\), mask threshold \(\mu=4\), learnable layers \(\lambda=10\)
        \State Compute singular defect direction $\nu_i$ for each layer $i$ of the pre-trained network
        \While {more than \(\rho\) of recent \(M\) images are not clear}
        \State Sample an image
        \ForAll{layers $i$}
        \State Find defective tokens of layer $i$ using $\nu_i$\Comment{See Equation~\eqref{eq:equation5}}
        \If{the number of defective tokens is less than \(\sigma\)}
        \State \textbf{continue}\Comment{Skip the current layer}
        \EndIf{}
        \State Compute loss using Equation~\eqref{eq:loss}
        \State Backward and update the parameters from layer $i-\lambda$ to layer $i$
        \State \textbf{break}\Comment{Skip remaining layers}
        \EndFor
        \EndWhile
    \end{algorithmic}
\end{algorithm}

\subsection{Loss Design}\label{sec:loss}
The core idea underlying our method is to first identify the defective tokens and then apply a spatial smoothness prior to regularizing them.
Let the patch tokens of a layer be $x_t$, $t=1,\dots, T$, where $T=H\times W$ is the number of tokens, and let the singular defect direction of the $i$th layer of the network be $\nu_i$.
We identify the defective tokens as follows.
First, define the logit $l_t$ as the absolute value of the inner product between the normalized patch token and $\nu_i$, \emph{i.e.},
\begin{equation}\label{eq:equation5}
    l_t = \left|\frac{x_t}{\|x_t\|}\cdot \nu_i \right|.
\end{equation}
Then, we take the set of defective tokens $\mathcal{D}$ to be those that deviate from the mean logit by more than the \emph{mask threshold} \(\mu=4\) times the standard deviation.

For a defective token \(x_t \in \mathcal{D}\), we define its learning target based on the weighted average of its \(3\times 3\) spatially neighboring tokens \(\mathcal{N}_t\).
Let token $x_{t'} \in \mathcal{N}_t$ be a neighboring token of \(x_t\). Then, we compute the coefficient
\begin{equation}
    c_{tt'} = \frac{\exp(-l_{t'}/\tau)}{\sum_{s\in \mathcal{N}_t} \exp(-l_s/\tau)},
\end{equation}
where $\tau$ is a temperature hyperparameter.
Additionally, we multiply $c_{tt'}$ with a $3\times 3$ Gaussian kernel and re-normalize the resulting coefficients. This step assigns greater weight to closer neighbors compared to farther ones. Finally, we utilize the resulting coefficients, denoted as $\tilde{c}_{tt'}$, to linearly combine the $3\times 3$ neighboring tokens into the learning target of each defective token $x_t$ as
\begin{equation}
    \tilde{x}_t = \sum_{t' \in \mathcal{N}_t} {\tilde{c}}_{tt'}x_{t'}.
\end{equation}
We define our loss function \(L\) as the average distance between the defective tokens and their respective learning targets, which can be expressed as
\begin{equation}\label{eq:loss}
    L = \frac{1}{|\mathcal{D}|}\sum_{t\in \mathcal{D}}\|x-\tilde{x}_t\|.
\end{equation}
If the number of defects for every layer is less than the skip threshold \(\sigma\), then we call this image \emph{clear}.

\subsection{Limiting the Number of Learnable Parameters}\label{sec:limit}
Given the fact that we fine-tune the model with significantly fewer images compared to the original training set, it becomes imperative to control the number of trainable parameters to avoid compromising the model's generalization ability in downstream tasks.
Our observation of a profound connection between the leading left singular vector of network operations and the empirical defect directions serves as the foundation for our approach.
Based on the intuition from the power method, the high norm of defects is related to the corresponding leading singular value. We thus propose to constrain learning to singular values only. Specifically, we decompose the weight of every linear layer in DINOv2 as \(USV^T\) using SVD and freeze the parameters \(U\) and \(V\) during fine-tuning.
This greatly reduces the number of learnable parameters.

Furthermore, our experiments reveal that further restricting the number of learnable parameters benefits feature quality preservation. Consequently, we opt to completely freeze most layers during fine-tuning. As illustrated in line 10 of Algorithm~\ref{alg:algorithm}, only the 10 layers preceding the first defective layer are trainable in each iteration. The effectiveness of this approach will be validated in Section~\ref{sec:ablation}.

\section{Experiments}
In this section, we first demonstrate the improvement resulting from our approach in the downstream task of unsupervised segmentation (Section~\ref{sec:unsupervised}).
We use two representative unsupervised segmentation methods, STEGO~\cite{hamilton2022unsupervised} and CAUSE~\cite{kim2023causal}.
The results of unsupervised segmentation demonstrate the importance of the repaired spatially smooth feature map in dense downstream tasks.
Then, we verify that our repaired DINOv2 retains feature quality.
To this end, we test classification performance on ImageNet-1K~\cite{deng2009imagenet}, which ensures the quality of the {\texttt{cls\_token}} (Section~\ref{sec:classification}), as well as supervised segmentation on ADE20k~\cite{zhou2017scene} and VOC2012~\cite{pascal-voc-2012} and depth estimation on NYUd~\cite{nyud}, which ensures the quality of the patch tokens (Section~\ref{sec:segmentation}).
Finally, we study our design choices and hyperparameter settings (Section~\ref{sec:ablation}).

In these experiments, we compare 3 models: The official release of DINOv2 giant, the DINOv2 giant model trained with registers~\cite{darcet2023vitneedreg}, and our repaired DINOv2 giant model based on the original DINOv2 without registers.

\subsubsection{Training Setting.}
We randomly select 30k images from the training set of ImageNet-1K to fine-tune for one epoch.
We use SGD with momentum \(0.9\) and weight decay 0.
The batch size is 1, and the learning rate is 0.005.
All input images are center-cropped and resized to \(518\times 518\).
The training procedure follows Algorithm~\ref{alg:algorithm}, which takes about six hours on a V100 GPU\@ for fine-tuning.
We limit the number of learnable layers to \(\lambda=10\) in each iteration.
The loss  Equation~\eqref{eq:loss} is computed on the first layer that has no less than \(\sigma=3\) defects.
Training stops if, during the latest \(M=500\) iterations, no loss was produced in more than \(1-\rho=75\%\) of the iterations.
The resulting network checkpoint is benchmarked on various datasets such as Cityscapes~\cite{Cordts2016Cityscapes}, Potsdam-3\cite{ji2019invariant}, VOC2012~\cite{pascal-voc-2012}, ADE20k~\cite{zhou2017scene}, \emph{etc.} in the following sections.

\subsection{Unsupervised Segmentation}\label{sec:unsupervised}

\begin{table}[tb]
  \caption{Results on unsupervised segmentation using STEGO\@.
    Backbones are frozen.
    The unsupervised results are shown in the Cluster columns.
    The Linear probe results are supervised and used for reference only.
  }\label{tab:stego}
  \centering
  \setlength{\tabcolsep}{0.4em}
  \resizebox{0.95\textwidth}{!}{
    \begin{tabular}{lcccccccc}
      \toprule
      \multirow{3}{*}{\shortstack[c]{~                                                                                                                                                                                                            \\~\\~\\Backbone for\\STEGO}} & \multicolumn{4}{c}{Cityscapes} & \multicolumn{4}{c}{Potsdam-3}                                                                                                                                                \\
      \cmidrule(r){2-5}\cmidrule(l){6-9}
                      & \multicolumn{2}{c}{\textbf{Cluster}} & \multicolumn{2}{c}{Linear} & \multicolumn{2}{c}{\textbf{Cluster}} & \multicolumn{2}{c}{Linear}                                                                                     \\
      \cmidrule(r){2-3}\cmidrule(lr){4-5}\cmidrule(lr){6-7}\cmidrule(l){8-9}
                      & mIoU                                 & Acc                        & mIoU                                 & Acc                        & mIoU               & Acc                & mIoU               & Acc                \\
      \midrule
      DINOv2          & \(19.38\)                            & \(72.54\)                  & \(43.00\)                            & \(91.69\)                  & \(67.01\)          & \(80.52\)          & \(76.47\)          & \(86.72\)          \\
      DINOv2-Register & \(18.62\)                            & \(67.00\)                  & \(\mathbf{43.97}\)                   & \(91.66\)                  & \(61.03\)          & \(75.69\)          & \(\mathbf{80.03}\) & \(\mathbf{88.97}\) \\
      DINOv2-SINDER   & \(\mathbf{21.77}\)                   & \(\mathbf{77.39}\)         & \(43.06\)                            & \(\mathbf{92.05}\)         & \(\mathbf{70.26}\) & \(\mathbf{82.40}\) & \(77.39\)          & \(87.31\)          \\
      \bottomrule
    \end{tabular}
  }
\end{table}

\begin{table}[tb]
  \caption{Results on unsupervised segmentation using CAUSE\@.
    Backbones are frozen.
  }\label{tab:cause}
  \centering
  \setlength{\tabcolsep}{0.1em}
  \begin{tabular}{lcccccccccccc}
    \toprule
    \multirow{3}{*}{\shortstack[c]{~                                                                                                                                                                                                                                                                                  \\~\\~\\Backbone for                                                                                                                                                                                                                                                                        \\CAUSE}} & \multicolumn{6}{c}{Cityscapes}  & \multicolumn{6}{c}{VOC2012}                                                                                                                                                                                                        \\
    \cmidrule(r){2-7}\cmidrule(l){8-13}
                    & \multicolumn{3}{c}{Without CRF} & \multicolumn{3}{c}{With CRF} & \multicolumn{3}{c}{Without CRF} & \multicolumn{3}{c}{With CRF}                                                                                                                                                                 \\
    \cmidrule(r){2-4}\cmidrule(lr){5-7}\cmidrule(lr){8-10}\cmidrule(l){11-13}

                    & mIoU                            & mAP                          & Acc                             & mIoU                         & mAP               & Acc               & mIoU              & mAP               & Acc               & mIoU              & mAP               & Acc               \\
    \midrule
    DINOv2          & \(31.4\)                        & \(45.2\)                     & \(85.2\)                        & \(31.5\)                     & \(57.6\)          & \(89.8\)          & \(55.8\)          & \(71.3\)          & \(91.7\)          & \(57.5\)          & \(79.0\)          & \(93.1\)          \\
    DINOv2-Register & \(33.3\)                        & \(51.2\)                     & \(87.6\)                        & \(35.3\)                     & \(71.6\)          & \(\mathbf{90.7}\) & \(48.9\)          & \(74.8\)          & \(90.9\)          & \(51.1\)          & \(78.8\)          & \(92.0\)          \\
    DINOv2-SINDER   & \(\mathbf{35.6}\)               & \(\mathbf{54.6}\)            & \(\mathbf{88.4}\)               & \(\mathbf{35.9}\)            & \(\mathbf{72.9}\) & \(\mathbf{90.7}\) & \(\mathbf{62.9}\) & \(\mathbf{85.6}\) & \(\mathbf{93.6}\) & \(\mathbf{63.8}\) & \(\mathbf{88.3}\) & \(\mathbf{94.1}\) \\
    \bottomrule
  \end{tabular}
\end{table}

As shown by the PCA visualization in Figure~\ref{fig:high_norm}, the advantage of the repair is a spatially smooth feature map.
Speculatively, our repaired DINOv2 can thus benefit dense prediction tasks such as unsupervised segmentation because the new feature map has clearer boundaries and more coherent semantics.
To verify this intuition, we compare our repaired DINOv2 with the original DINOv2 and DINOv2-Register using two representative unsupervised segmentation methods, namely, STEGO and CAUSE\@.
We follow the training settings and the processing of benchmark datasets in their respective papers.
Detailed hyper-parameters and configurations can be found in the Appendix.
From Table~\ref{tab:stego}, we observe that, compared with DINOv2, our DINOv2-SINDER improves the mIoU of STEGO on Cityscapes~\cite{Cordts2016Cityscapes} by +2.39\%, and on Potsdam-3\cite{ji2019invariant} by +3.25\% in the unsupervised cluster setting.
The performance of the supervised linear setting is used for reference only.
From Table~\ref{tab:cause}, we observe that, compared with DINOv2, our DINOv2-SINDER improves the mIoU of CAUSE on Cityscapes by +4.2\% and +4.4\% in the without/with CRF settings, respectively.
On the VOC2012~\cite{pascal-voc-2012} dataset, the improvements are +7.1\% and +6.3\%, respectively.
The tables also show that our DINOv2-SINDER outperforms the DINOv2-Register.
These results confirm that our proposed SINDER is effective on the dense downstream task of unsupervised segmentation.

\subsection{Classification}\label{sec:classification}

\begin{table}[tb]
  \caption{Results on ImageNet-1K classification. Backbones are frozen.
  }\label{tab:cls}
  \centering
  \setlength{\tabcolsep}{0.5em}
  \begin{tabular}{lcccccc}
    \toprule
    \multirow{2}{*}{\shortstack[c]{~                                                                    \\~\\Backbone}} & \multicolumn{2}{c}{KNN} & \multicolumn{2}{c}{Linear}                         \\
    \cmidrule(r){2-3}\cmidrule(l){4-5}

                    & Top1               & Top5               & Top1               & Top5               \\
    \midrule
    DINOv2          & \(83.53\)          & \(94.01\)          & \(86.53\)          & \(97.65\)          \\
    DINOv2-Register & \(\mathbf{83.69}\) & \(93.12\)          & \(\mathbf{87.10}\) & \(\mathbf{97.95}\) \\
    DINOv2-SINDER   & \(83.51\)          & \(\mathbf{94.15}\) & \(86.29\)          & \(97.61\)          \\
    \bottomrule
  \end{tabular}
\end{table}

We test the classification performance of our repaired DINOv2-SINDER on ImageNet-1K.
We follow the evaluation protocol of~\cite{oquab2023dinov2}.
Specifically, we test KNN and linear probe on frozen backbones.
The top-1 and top-5 accuracies of the three compared models are provided in Table~\ref{tab:cls}.
The top-1 and top-5 accuracies of DINOv2-SINDER are on par with the original DINOv2, for both KNN and linear probe.
Compared with DINOv2-Register, the top-1 accuracy of KNN is -0.18\% lower, but the top-5 accuracy of KNN is +1.03\% higher.
The top-1 and top-5 accuracies for the linear probe are -0.81\% and -0.34\% lower than the DINOv2-Register, which is similar to those of the original DINOv2.
Note that DINOv2-Register requires full retraining from scratch, whereas our fine-tuning uses substantially fewer resources.
This comparison validates that our fine-tuned model maintains the feature quality of the \texttt{cls\_token}.

\subsection{Supervised Segmentation}\label{sec:segmentation}

\begin{table}[tb]
  \caption{Results on supervised segmentation.
    Backbones are frozen.
  }\label{tab:sseg}
  \centering
  \setlength{\tabcolsep}{0.48em}
  \begin{tabular}{lcccccccc}
    \toprule
    \multirow{3}{*}{\shortstack[c]{~                                                                                                                                                                                                \\~\\~\\~\\Backbone}} & \multicolumn{4}{c}{ADE20k} & \multicolumn{4}{c}{VOC2012}                                                                                                                                                      \\
    \cmidrule(r){2-5}\cmidrule(l){6-9}

                    & \multicolumn{2}{c}{Linear} & \multicolumn{2}{c}{Multiscale} & \multicolumn{2}{c}{Linear} & \multicolumn{2}{c}{Multiscale}                                                                                     \\
    \cmidrule(r){2-3}\cmidrule(lr){4-5}\cmidrule(lr){6-7}\cmidrule(l){8-9}

                    & mIoU                       & aAcc                           & mIoU                       & aAcc                           & mIoU               & aAcc               & mIoU               & aAcc               \\
    \midrule
    DINOv2          & \(48.83\)                  & \(81.46\)                      & \(53.24\)                  & \(84.00\)                      & \(83.05\)          & \(96.17\)          & \(86.01\)          & \(97.01\)          \\
    DINOv2-Register & \(49.03\)                  & \(81.09\)                      & \(53.62\)                  & \(83.90\)                      & \(83.27\)          & \(96.15\)          & \(86.54\)          & \(97.12\)          \\
    DINOv2-SINDER   & \(\mathbf{51.11}\)         & \(\mathbf{82.70}\)             & \(\mathbf{54.78}\)         & \(\mathbf{84.75}\)             & \(\mathbf{84.63}\) & \(\mathbf{96.57}\) & \(\mathbf{86.94}\) & \(\mathbf{97.25}\) \\
    \bottomrule
  \end{tabular}
\end{table}

To verify that the feature quality of the patch tokens is at least equally good, we perform supervised segmentation on ADE20k and VOC2012 using the linear probe with frozen backbones.
Two training settings are tested.
The linear setting only uses the last feature map, while the multi-scale setting uses the feature maps of the last four layers.
The results are provided in Table~\ref{tab:sseg}.
Compared with the original DINOv2, our repaired version improves the mIoU by +2.28\% and +1.54\%, respectively, for the linear and multi-scale settings on ADE20k, and +1.58\% and +0.93\% on VOC2012.
This shows the superiority of our method.
Compared with DINOv2-Register, our DINOv2-SINDER improves the mIoU by +2.08\% and +1.16\% on ADE20k for the linear and multi-scale settings respectively, and +1.36\% and +0.40\% on VOC2012.
This is surprising considering that DINOv2-Register has mitigated the high-norm defects at the cost of full retraining.
Although DINOv2-Register is not as performant as our DINOv2-SINDER, it is still better than DINOv2.
This comparison demonstrates that our method not only retains the quality of the patch tokens but also improves the dense prediction downstream task in the supervised setting.

\subsection{Depth Estimation}\label{sec:depth}

\begin{table}[tb]
  \caption{Results on NYUd depth estimation.
    Backbones are frozen.
  }\label{tab:depth}
  \centering
  \setlength{\tabcolsep}{0.5em}
  \begin{tabular}{lcccccc}
    \toprule
    Backbone        & Linear 1           & Linear 4           & DPT                \\
    \midrule
    DINOv2          & \(0.370\)          & \(0.309\)          & \(0.242\)          \\
    DINOv2-Register & \(0.367\)          & \(0.302\)          & \(\mathbf{0.234}\) \\
    DINOv2-SINDER   & \(\mathbf{0.337}\) & \(\mathbf{0.294}\) & \(0.249\)          \\
    \bottomrule
  \end{tabular}
\end{table}

We evaluate the patch features using depth estimation on the NYU Depth v2 dataset, following the testing protocol in~\cite{oquab2023dinov2}.
There are three settings.
(1) \texttt{Linear~1} uses the last layer feature map from the frozen backbone and concatenates the \texttt{cls\_token} to patch tokens.
The feature map is bilinear resized to the original resolution of the input image.
The depth range is uniformly divided into 256 bins and a linear layer predicts which bin the pixel should belong to.
(2) \texttt{Linear~4} is similar to \texttt{Linear~1}, except that it concatenates the tokens from layers \(9, 19, 29, 39\).
(3) \texttt{DPT} uses the DPT decoder on features of the frozen backbone. Regression losses are used in the setting.
The results are shown in Table~\ref{tab:depth}.
We see that SINDER outperforms DINOv2 and DINOv2-Register in the two linear settings, showing that removing defective tokens has a greater benefit for simple head structures.
For complicated head DPT, the performance is on par with DINOv2 and slightly worse than DINOv2-Register.

\subsection{Ablation Study}\label{sec:ablation}
\subsubsection{Constrained Parameter Fine-tuning.}
To repair the singular defects while keeping feature quality, we need to strictly constrain the freedom of parameter learning.
To show the importance of constrained parameter fine-tuning, we compare five settings with gradually stronger constraints.
1. Learning the singular values of the weight matrices together with the biases of all linear layers.
2. Only learning the singular values of the weight matrices of the linear layers, except for the query matrices and K matrices in the attention.
3. Further constraining the learnable layers in the second setting to 15 layers in each iteration.
4. Constraining the learnable layers to 10 in each iteration.
5. Constraining the learnable layers to 5 in each iteration.
The results are shown in Table~\ref{tab:ablation1}.
A general trend is that the fewer learnable parameters, the more the classification accuracy and segmentation performance are preserved.
However, in the extreme case of too few learnable parameters, there is no room left for improvement in segmentation.
According to this study, we choose the balanced setting of restricting 10 layers.

\begin{table}[tb]
  \caption{Constrained parameter fine-tuning with gradually stronger constraints.
  }\label{tab:ablation1}
  \centering
  \setlength{\tabcolsep}{0.65em}
  \begin{tabular}{lcccc}
    \toprule
    \multirow{2}{*}{\shortstack[c]{~                                                                                          \\~\\Setting}}              & \multicolumn{2}{c}{KNN (ImageNet)} & \multicolumn{2}{c}{Seg. (ADE20k)}                                           \\
    \cmidrule(r){2-3}\cmidrule(l){4-5}

                                          & Top1               & Top5               & mIoU               & aAcc               \\
    \midrule
    Singular Value and Bias               & \(6.64\)           & \(16.03\)          & \(13.77\)          & \(61.91\)          \\
    Singular Value except QK              & \(80.12\)          & \(92.82\)          & \(45.53\)          & \(80.62\)          \\
    Singular Value except QK in 15 Layers & \(82.81\)          & \(92.88\)          & \(49.85\)          & \(82.51\)          \\
    Singular Value except QK in 10 Layers & \(83.51\)          & \(\mathbf{94.15}\) & \(\mathbf{51.11}\) & \(\mathbf{82.70}\) \\
    Singular Value except QK in 5 Layers  & \(\mathbf{83.53}\) & \(93.15\)          & \(50.61\)          & \(82.65\)          \\
    \bottomrule
  \end{tabular}
\end{table}

\begin{table}[tb]
  \caption{Comparison of different values of skip threshold \(\sigma\) and mask threshold \(\mu\).
  }\label{tab:ablation2}
  \centering
  \setlength{\tabcolsep}{0.65em}
  \def\arraystretch{0.96}
  \resizebox{0.75\textwidth}{!}{
    \begin{tabular}{lcccc}
      \toprule
      \multirow{2}{*}{\shortstack[c]{~                                                                                \\~\\Setting}}    & \multicolumn{2}{c}{KNN (ImageNet)} & \multicolumn{2}{c}{Seg. (ADE20k)}                                                                \\
      \cmidrule(r){2-3}\cmidrule(l){4-5}

                                  & Top1               & Top5               & mIoU               & aAcc               \\
      \midrule
      Skip Less than \(\sigma=0\) & \(83.33\)          & \(94.12\)          & \(\mathbf{51.19}\) & \(\mathbf{82.80}\) \\
      Skip Less than \(\sigma=3\) & \(83.51\)          & \(\mathbf{94.15}\) & \(51.11\)          & \(82.70\)          \\
      Skip Less than \(\sigma=5\) & \(\mathbf{83.52}\) & \(93.11\)          & \(50.71\)          & \(82.53\)          \\
      \midrule
      Mask Threshold \(\mu=3.5\)  & \(83.33\)          & \(93.10\)          & \(50.93\)          & \(\mathbf{82.72}\) \\
      Mask Threshold \(\mu=4\)    & \(\mathbf{83.51}\) & \(\mathbf{94.15}\) & \(\mathbf{51.11}\) & \(82.70\)          \\
      Mask Threshold \(\mu=4.5\)  & \(83.50\)          & \(93.09\)          & \(50.50\)          & \(82.51\)          \\
      \bottomrule
    \end{tabular}
  }
\end{table}

\subsubsection{Dynamic Layer Loss.}

To decide which layer to apply the loss to, we experimented with different hyper-parameter values for the skip threshold \(\sigma\) and the logit mask threshold \(\mu\).
The results are shown in Table~\ref{tab:ablation2}.
A general trend is that if it is easier to skip layers, then stronger KNN performance is preserved.
However, the improvement in segmentation is then limited.
This is because more skipped layers cause earlier termination according to line 2 of Algorithm~\ref{alg:algorithm}.
For the mask threshold, we find that the value \(\mu=4\) works well for DINOv2.

\begin{figure}[tb]
  \centering
  \includegraphics[height=6.4cm]{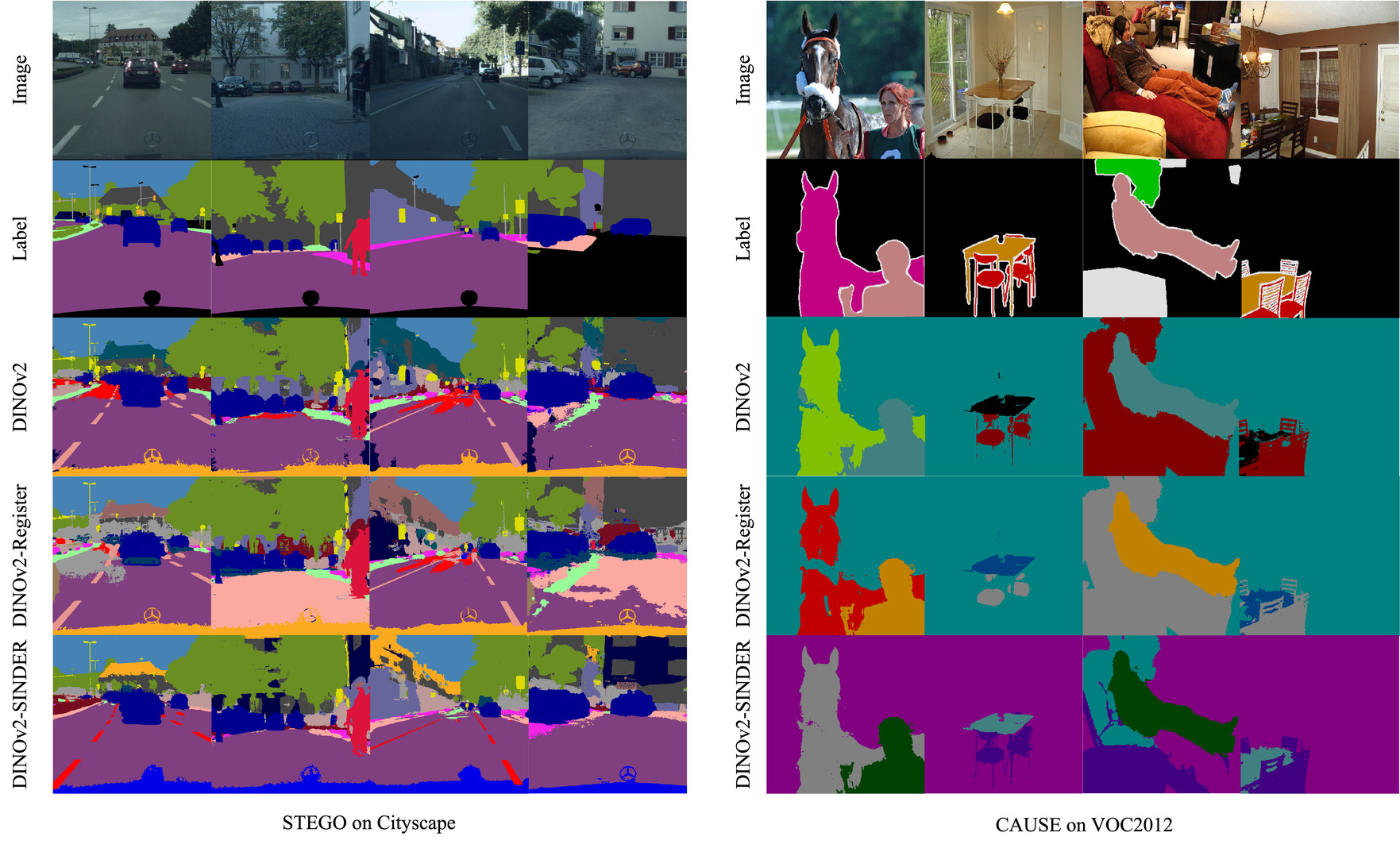}
  \caption{Visualization of unsupervised segmentation on Cityscapes using STEGO\@.
  }\label{fig:visualization}
\end{figure}

\section{Related Work}

\subsubsection{Self-supervised Models.}
The recent surge in SSL methodologies began with the application of the contrastive loss~\cite{chen2020simple} or the cosine loss~\cite{grill2020bootstrap} on Siamese convolutional neural networks.
Owing to their multi-modality friendly nature, Vision Transformers (ViTs)~\cite{dosovitskiy2020vit} are swiftly replacing CNNs as the mainstream backbone.
Leveraging their global attention property, numerous SSL methods have been proposed for pre-training these networks.
For instance, MoCov3~\cite{chenempirical} follows the contrastive setting, MAE~\cite{he2022masked} reconstructs masked patch tokens, while iBoT~\cite{zhou2021ibot} and I-JPEA~\cite{assran2023self} learn to predict feature vectors of masked or nearby regions.
Despite achieving promising performance across various downstream tasks, DINOv2~\cite{oquab2023dinov2} advances this direction further by combining the advantages of prior arts such as iBoT and DINO~\cite{caron2021emerging} in the loss, improving the data curation, and adopting other dedicated engineering efforts.
Trained on the large-scale dataset LVD-142M~\cite{oquab2023dinov2}, DINOv2 exhibits impressive performance and robust zero-shot ability, heralding a new era for training foundational vision models.

\subsubsection{Analyzing Self-supervised Models and Transformers.}

Given the significance of SSL in various applications, researchers have dedicated efforts to understand its mechanisms. Some studies~\cite{wang2022chaos,wang2020understanding} interpret contrastive learning by dissecting the loss function into several interpretable terms, while others analyze SSL from an augmentation perspective~\cite{xiao2020should,zhang2022leverage,tian2020makes}.
Recently, it has been observed that the object features of iBoT and I-JPEA exhibit coupling~\cite{qiu2023mind}, thereby impeding their ability to distinguish different objects. Similarly, DINOv2 has been found to possess defective tokens~\cite{darcet2023vitneedreg}, which undermine its performance on dense prediction downstream tasks.
While proposing re-training ViTs with more tokens, these studies do not explain the underlying phenomenon.

The pairwise positive relationships between training samples using spectral methods are investigated in~\cite{balestriero2022contrastive}. Notably, they utilize Singular Value Decomposition (SVD) in their analysis, although the decomposition is applied to the representation matrix composed of feature maps. Our analysis diverges from theirs as our SVD is applied to the network parameters rather than the features themselves. The work of~\cite{geshkovski2023mathematical} analyzes transformers based on their interpretation as interacting particle systems. Specifically, they observe the emergence of clusters over time. However, their analysis is limited to a simplified ideal transformer architecture, which disregards the MLP block and multi-head attention. Moreover, their focus is primarily on the case where \(Q=K=V=I_d\), which is not realistic. The work of~\cite{geshkovski2023emergence} assumes a fixed \(V\), which is restrictive as it does not account for the varying semantics learned across different layers. By contrast, our analysis directly examines the pre-trained weights, including the learned parameters of  \(Q\), \(K\), and \(V\). Furthermore, we consider the multi-head structure as well as the MLP block in our analysis.

\section{Limitation and Social Impact}
This work focuses on repairing existing pre-trained networks.
How to avoid singular defects from training is left for future work.
We primarily focus on the study of DINOv2, and we hope our treatment could motivate more research on the understanding of more transformer-based networks such as GPTs.
Our method of repairing existing networks requires substantially fewer computation resources and data consumption, which reduces the carbon emissions and human labor in curating data, compared to the existing approach of fully retraining.

\section{Conclusion}
In this paper, we have introduced a principled way to connect the high-norm defective tokens in DINOv2 with the leading left singular vector of the pre-trained weights.
Based on this finding, we propose to repair DINOv2 by fine-tuning using a smooth prior loss optimized on a restricted number of parameters.
Our experiments have shown that our singular defect direction prediction aligns well with the empirical defect direction, and our repaired DINOv2 improves unsupervised pixel-level prediction downstream tasks while retaining feature quality.

\section*{Acknowledgements}
This work was supported in part by the Swiss National Science Foundation via the Sinergia grant CRSII5-180359.
\appendix

\section{Details on PCA Visualization}
Our PCA visualization in Figures 1 and 3 follows the procedure in~\cite{oquab2023dinov2}.
Given the feature map of an image, containing \(H\times W\) patch tokens, we first extract the three leading principal components of these tokens.
Each token will become a dim-3 vector after the PCA.
Then we scale each component to the range \(0-255\) and interpret them as the RGB channels.
The tokens are reshaped to resolution \(H\times W\), and we get the resulting PCA visualization.

\section{Visualization of Clamping Singular Values}
\begin{figure}[h]
  \centering
  \includegraphics[width=0.99\linewidth]{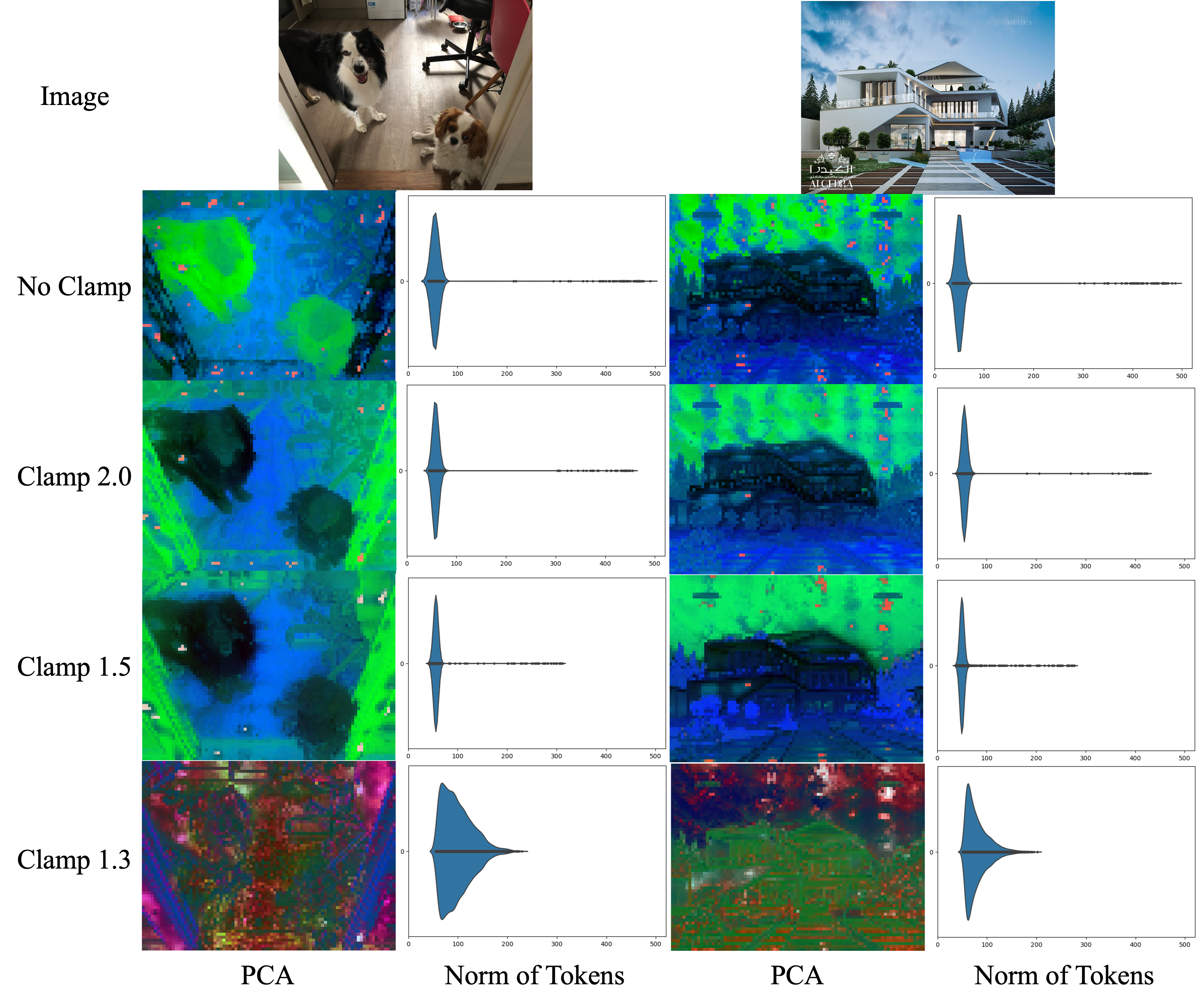}
  \caption{Visualization after clamping the singular values of linear layers.
    The results of the two images are illustrated.
    The first and third columns are the PCA visualization of the feature map in the last layer.
    The second and fourth columns are the violin plots of the norm of the corresponding tokens.
  }\label{fig:clamp}
\end{figure}

As a sanity check in Section 2, we clamp the singular value of the weights of linear layers in DINOv2 to a smaller value.
The PCA visualization is shown in Figure~\ref{fig:clamp}.
For each linear layer, we decompose the weight matrix using SVD (for example, \(W=USV^T\)) and then clamp the singular values \(S\) to be less than a threshold \(\gamma\) and get \(\tilde{S} \).
At last, we replace the weight matrix with \(U\tilde{S}V\).
In Figure~\ref{fig:clamp}, we compared with \(\gamma=2.0\), \(1.5\), and \(1.3\), respectively.
We can see that as \(\gamma\) decreases, the norms of defective patches also decrease, and the number of defective tokens becomes less.
However, when \(gamma\) is too small, the semantics of the feature maps seem corrupted.
So we would prefer learned optimal singular values rather than trimming them according to some manually designed thresholds.

\section{Visualization of Learning Target}
We visualize the learning target defined by Equation (7) in the last row of Figure~\ref{fig:violin_defect}.
The feature maps of the 9th, 19th, 29th, and 39th layers for the villa image are illustrated.
For demonstrative purposes, we show the learning targets for all pixels, which are visually smooth.
However, in real fine-tuning, only pixels that are detected as defective contribute to the loss.
The corresponding singular defects using mask threshold \(\mu=4\) are shown in the third row in Figure~\ref{fig:violin_defect}.

\section{Visualization of Angles Between \(\nu_i\) and Patch Tokens}

\begin{figure}[tb]
  \centering
  \includegraphics[width=0.98\linewidth]{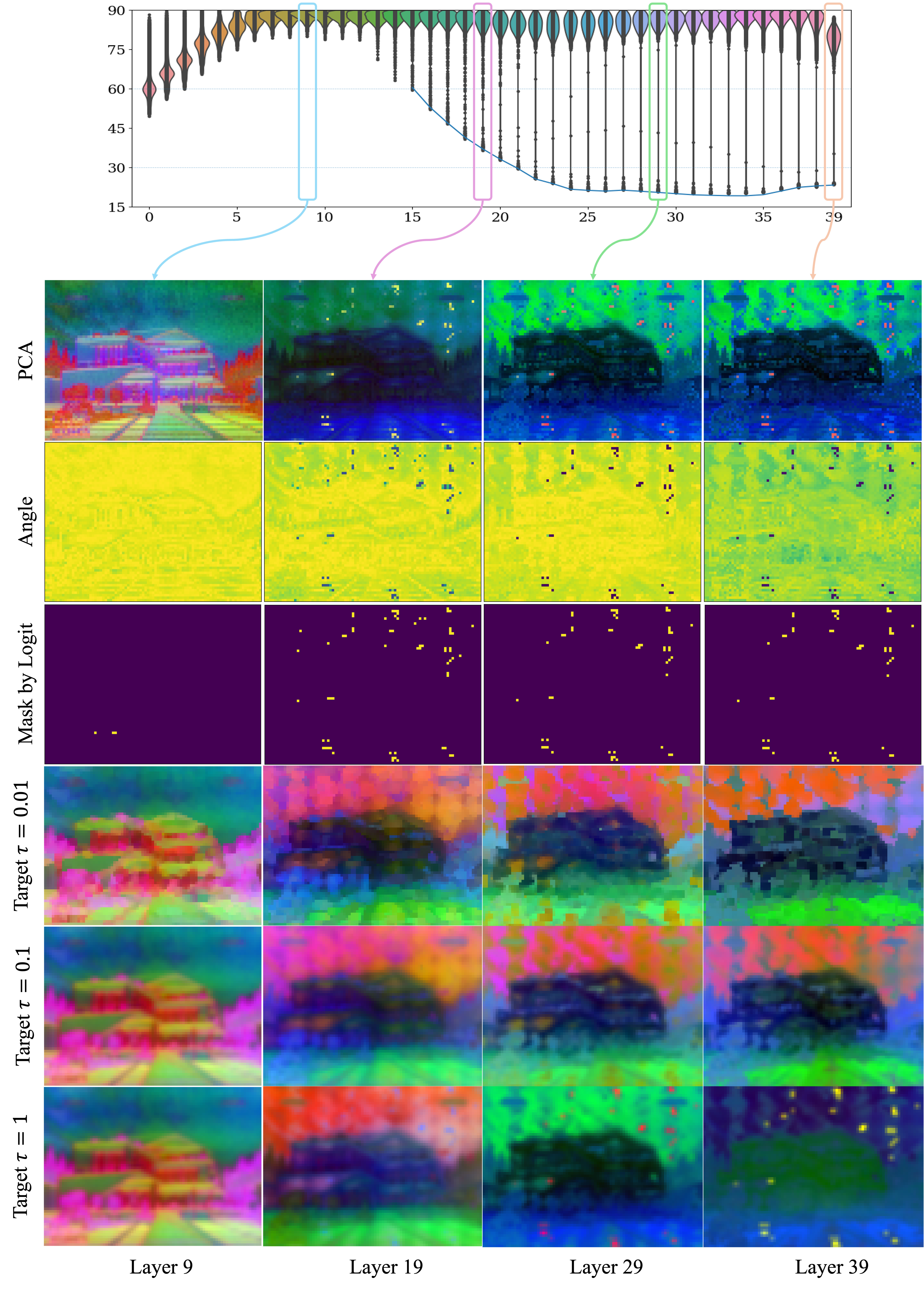}
  \caption{The violin plot is the visualization of angles between the theoretical singular defect direction \(\nu_i\) and patch tokens.
    The first row below the violin plot is the PCA visualization of patch tokens in the 9th, 19th, 29th, and 39th layers.
    The second row is the heat map of the angle between \(\nu_i\) and patch tokens.
    The darker the color, the smaller the angles.
    The third row is the defective tokens detected by logits defined in Equation (5).
    The last three rows are the learning target under the temperature hyper-parameter \(\tau=0.01, 0.1, 1\).
    We use \(\tau=0.1\) in our experiments.
  }\label{fig:violin_defect}
\end{figure}

In Figure~2d, the violin plot of the angles between the theoretical singular defect directions \(\nu_i\) for layer-\(i\) and the patch tokens of the villa image are illustrated.
To show that the isolated, anomalous points in the violin plot are indeed defective tokens, we present the corresponding PCA visualizations and the heatmap of angles in the first and second rows of Figure~\ref{fig:violin_defect}.
In the angles heatmap, darker pixels mean that the angles between \(\nu_i\) and the patch tokens are smaller.

\section{Visualization of the Learned Singular Values}

\begin{figure}[t]
  \centering
  \begin{subfigure}{0.89\linewidth}
    \includegraphics[width=0.99\linewidth]{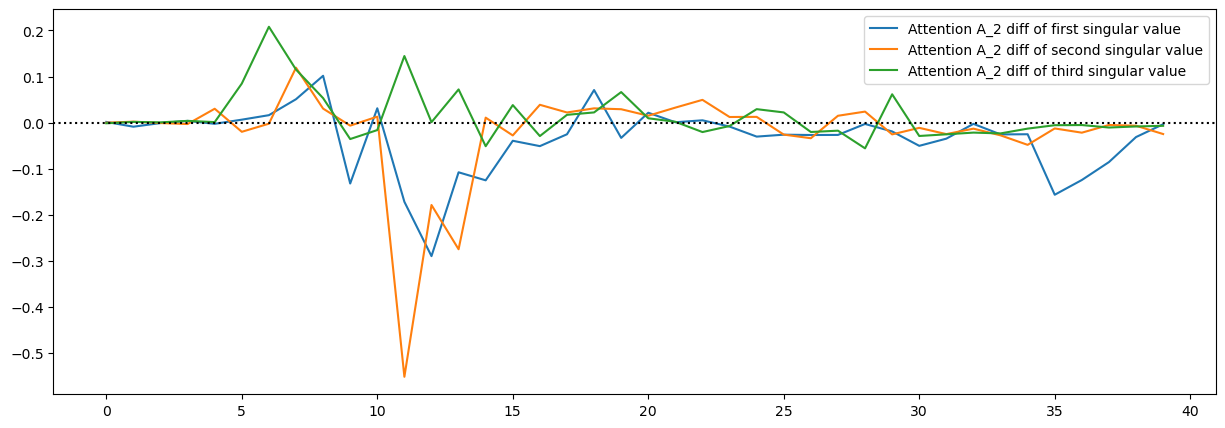}
    \caption{Diff of Singular Values of \(A_2\) in the Attention Block}
  \end{subfigure}
  \begin{subfigure}{0.89\linewidth}
    \includegraphics[width=0.99\linewidth]{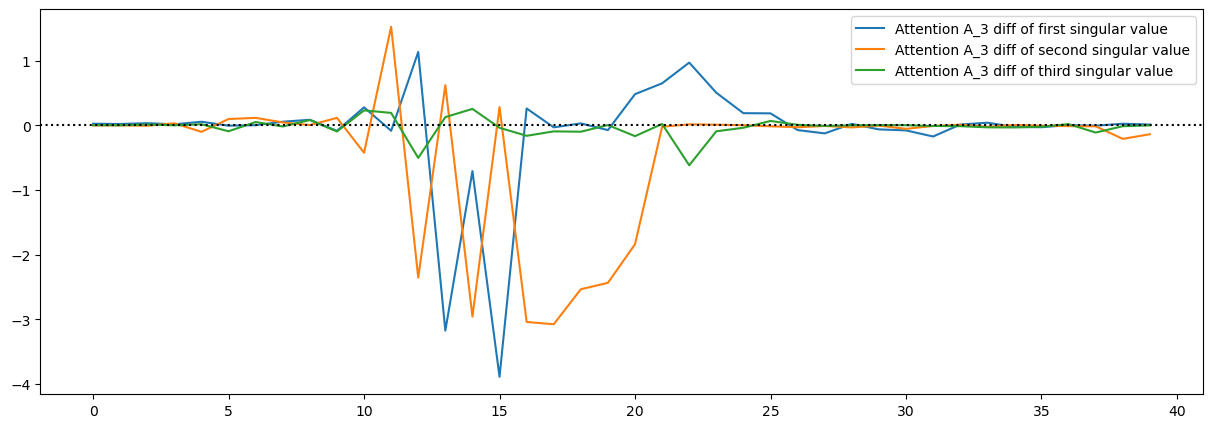}
    \caption{Diff of Singular Values of \(A_3\) in the Attention Block}
  \end{subfigure}
  \caption{
    Difference between the learned singular values and the original singular values.
    Changes in the remaining singular values are relatively small compared to the leading ones.
    So, only the first three leading singular values are shown to avoid cluttering the figure.
    The \(x\)-axis is the layer index.
  }\label{fig:appendix2}
\end{figure}

We show the difference between the learned singular values and the original singular values in Figure~\ref{fig:appendix2}.
We observe that the differences are more striking for layers 5--25, and changes in other layers are modest.
Generally speaking, the learned singular values are smaller than the original values.

\begin{figure}[tb]
  \centering
  \includegraphics[width=0.98\linewidth]{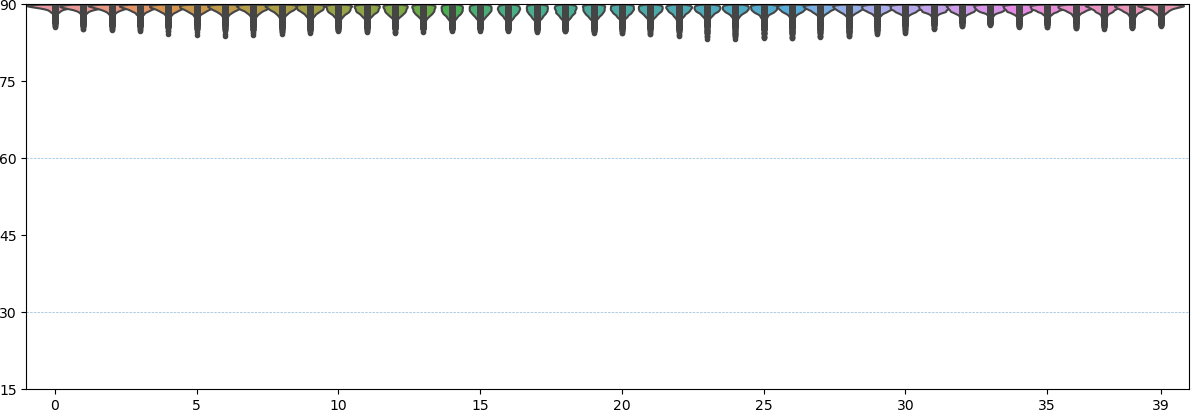}
  \caption{The violin plot of angles between a random direction and the patch tokens of the villa image.
    The \(x\)-axis is the layer index.
  }\label{fig:violin_randdir}
\end{figure}
\FloatBarrier

\section{Visualization of Angles between Random Direction and Patch Tokens}
To complement Figure 2, we show the violin plot of angles between a random direction and the patch tokens of the villa image in Figure~\ref{fig:violin_randdir}.
We can see that the pattern of the violin plot using a random direction is drastically different from the violin plots in Figure 2, which uses the leading left singular vectors.
This shows that a random direction cannot be used to detect defective tokens.

\section{Detailed Configurations of Experiments}
\subsection{STEGO}
\subsubsection{Dataset}
We follow STEGO~\cite{hamilton2022unsupervised} to process the datasets, specifically, 27 classes of Cityscapes~\cite{Cordts2016Cityscapes}, and 3 classes of Potsdam-3~\cite{ji2019invariant} are evaluated.
We resize the images to \(392\times 392\) with center crop in training and \(560\times 560\) in testing.
The training images are five-cropped.
\subsubsection{Hyper-parameter}
\begin{table}[tb]
  \caption{Hyper-parameters of STEGO.
  }\label{tab:param_stego}
  \centering
  \begin{tabular}{llcccccc}
    \toprule
    {\shortstack[c]{Dataset                                                                        \\~}} & {\shortstack[c]{Backbone\\~}}              & {\shortstack[c]{neg inter \\weight}}    & {\shortstack[c]{pos inter \\ weight}} &  {\shortstack[c]{pos intra \\weight}} &   {\shortstack[c]{neg inter \\shift }}& {\shortstack[c]{pos inter \\shift }} &{\shortstack[c]{ pos intra \\shift  }}   \\
    \midrule
    Cityscapes & DINOv2          & \(0.90\) & \(0.60\) & \(1.00\) & \(0.30\) & \(0.20\) & \(0.45\) \\
    Cityscapes & DINOv2-Register & \(0.80\) & \(0.65\) & \(0.90\) & \(0.30\) & \(0.20\) & \(0.45\) \\
    Cityscapes & DINOv2-SINDER   & \(0.80\) & \(0.65\) & \(0.90\) & \(0.30\) & \(0.45\) & \(0.60\) \\
    Potsdam-3  & DINOv2          & \(0.90\) & \(0.60\) & \(1.00\) & \(0.30\) & \(0.20\) & \(0.45\) \\
    Potsdam-3  & DINOv2-Register & \(0.90\) & \(0.60\) & \(1.00\) & \(0.30\) & \(0.20\) & \(0.45\) \\
    Potsdam-3  & DINOv2-SINDER   & \(0.90\) & \(0.60\) & \(1.00\) & \(0.40\) & \(0.45\) & \(0.45\) \\
    \bottomrule
  \end{tabular}
\end{table}

We extended the STEGO's official codebase to support DINOv2 backbones.
We use the hyper-parameters in STEGO's official repository, except for the backbone- and dataset-sensitive parameters, which are listed in Table~\ref{tab:param_stego}.
The hyper-parameters of STEGO used in the results of Table~1 are listed in Table~\ref{tab:param_stego}.

\subsection{CAUSE}
\subsubsection{Dataset}
We follow CAUSE~\cite{kim2023causal} to process the datasets, specifically, 27 classes of Cityscapes, and 21 classes of PASCAL-VOC~\cite{pascal-voc-2012} are evaluated.
Both training and testing resolution are \(322\times 322\).
\subsubsection{Hyper-parameter}
We use the official codebase in CAUSE and adopt the default settings for all our experiments.
specifically, we use the setting of CAUSE-TR\@.

\subsection{Classification KNN}
We use the KNN implementation in the official codebase of DINOv2\@.
The ImageNet-1K~\cite{deng2009imagenet} dataset is used.
The KNN performance on the validation set has been reported in Table 3.
Specifically, \(K=10, 20, 100, 200\) are tested and the setting with the best top1 was reported.

\subsection{Classification Linear Probe}
We follow the linear probe implementation in the official codebase of DINOv2\@.
The ImageNet-1K dataset is used.
The linear probe performance on the validation set is reported in Table 3.
The linear layer was trained for 10 epochs under learning rates of 1e-5, 2e-5, 5e-5, 0.0001, 0.0002, 0.0005, 0.001, 0.002, 0.005, 0.01, 0.02, 0.05, and 0.1, respectively, and the setting with the best top1 was reported.

\subsection{Supervised Segmentation}
We follow the evaluation protocol for supervised segmentation in DINOv2 and implement the training and testing using mmsegmentation~\cite{mmseg2020}.
Specifically, a linear layer is trained to predict classes from patch tokens.
In the Linear setting, both training and testing images are resized to \(512\times 512\).
For the Multiscale setting, they are rescaled to \(640\times 640\).
Moreover, for the Multiscale setting, the patch tokens of the last four layers are concatenated, and the multiscale test-time augmentation was used in testing.
For both ADE20k~\cite{zhou2017scene} and VOC2012, 40,000 iterations were trained.

\bibliographystyle{splncs04}
\bibliography{main}
\end{document}